\def\year{2020}\relax
\documentclass[letterpaper]{article} 
\usepackage{aaai20}  
\usepackage{times}  
\usepackage{helvet} 
\usepackage{courier}  
\usepackage[hyphens]{url}  
\usepackage{graphicx} 
\usepackage{graphicx}  
\usepackage{amssymb}
\usepackage{bm}
\usepackage{amsthm}
\usepackage{amsmath}

\DeclareMathOperator*{\argmin}{arg\,min}
\usepackage{makecell}
\usepackage{mathtools}
\usepackage{booktabs}
\usepackage{multirow}
\usepackage[ruled,linesnumbered]{algorithm2e}
\usepackage{xcolor}
\usepackage{floatrow}
\newfloatcommand{capbtabbox}{table}[][\FBwidth]
\title{Channel Pruning Guided by Classification Loss and Feature Importance}
\author{
Jinyang Guo\textsuperscript{\rm 1},
Wanli Ouyang\textsuperscript{\rm 2},
Dong Xu\textsuperscript{\rm 1}\\
\textsuperscript{\rm 1}School of Electrical and Information Engineering, The University of Sydney \\
\textsuperscript{\rm 2}The University of Sydney, SenseTime Computer Vision Research Group, Australia\\
\{jinyang.guo, wanli.ouyang, dong.xu\}@sydney.edu.au
}
 
\begin{document}

\maketitle

\begin{abstract}
In this work, we propose a new layer-by-layer channel pruning method called Channel Pruning guided by classification Loss and feature Importance (CPLI). In contrast to the existing layer-by-layer channel pruning approaches that only consider how to reconstruct the features from the next layer, our approach additionally take the classification loss into account in the channel pruning process. We also observe that some reconstructed features will be removed at the next pruning stage. So it is unnecessary to reconstruct these features. To this end, we propose a new strategy to suppress the influence of unimportant features (\textit{i.e.}, the features will be removed at the next pruning stage). Our comprehensive experiments on three benchmark datasets, \textit{i.e.}, CIFAR-10, ImageNet, and UCF-101, demonstrate the effectiveness of our CPLI method.
\end{abstract}

\section{Introduction}
\label{Sec:Intro}
While deep learning methods have achieved remarkable success in many computer vision tasks, it is still a challenging task to deploy convolutional neural networks (CNNs) on mobile devicies due to tight computational resources. Several model compression technologies (\textit{e.g.}, quantization and tensor factorization) were recently developed to improve the efficiency of the CNNs, among which channel pruning is one of the most popular techniques. 

The channel pruning approaches can be roughly categorized as loss minimization methods \cite{molchanov2017taylor,lecun1990optimal,molchanov2019importance} and layer-by-layer approaches~\cite{he2017channel,luoiccv2017}. The loss minimization methods iteratively remove the channels with least effect on the final loss across the entire network. However, in order to evaluate the effect of channels on the final loss, the fine-tuning process needs to be performed frequently in these methods, which makes the channel pruning process slow. On the other hand, the layer-by-layer methods select informative channels and adjust model parameters by minimizing the reconstruction error of the features in the next layer. These methods are much faster because they prune the channels within a layer in one step and the fine-tuning process is performed only once. However, the layer-by-layer methods do not consider the final loss. The channels selected in these methods may have little effect on the final loss, leading to a sub-optimal solution for channel selection. In addition, the existing layer-by-layer approaches treat the reconstruction of each feature equally important, which leads to the \textit{next-layer feature removal} problem. Specifically, at the current pruning stage, the layer-by-layer approaches will minimize the reconstruction error of a feature from the next layer under the assumption that this feature will be kept in the compressed model. However, this assumption will be violated if this feature will be removed at the next pruning stage, and reconstruction of this feature will thus become unnecessary.

To solve the aforementioned problems, we propose a new channel pruning approach called Channel Pruning guided by classification Loss and feature Importance (CPLI), in which we follow the layer-by-layer approaches for faster speed and take the final loss into account to solve the sub-optimal problem for channel selection. In order to address the \textit{next-layer feature removal} problem, we also propose a new strategy to pay more attention to important features (\textit{i.e.}, the features to be kept in the compressed model) and ignore unimportant features (\textit{i.e.}, the features to be removed at the next pruning stage) . Specifically, our newly proposed method selects the channels in each layer by solving a LASSO optimization problem based on two aspects: 1) the cross-entropy loss and 2) the importance of features. In order to avoid frequently performing the fine-tuning process, we adjust the weights by solving a least square optimization problem based on the features from the next layer, which has a close-form solution. Moreover, due to the close-form solution of the least square optimization problem, the adjusted weights in this process can converge better than the alternative approaches, where the weights are adjusted by using simple fine-tuning techniques. Therefore, the channel selection process in our method is based on well-learned weights, which is more accurate than the existing loss minimization approaches.

The main contributions of this work are as follows: First, our newly proposed method takes advantage of efficiency from the layer-by-layer channel pruning approaches while also taking the final loss into account, which can partially solve the sub-optimal problem for channel selection in the layer-by-layer channel pruning methods. Second, our approach can ignore reconstruction of unimportant features to address the \textit{next-layer feature removal} problem. As a result, our method can select more informative channels when compared with the existing layer-by-layer channel pruning approaches. Comprehensive experiments on three benchmark datasets demonstrate the effectiveness of our newly proposed approach for model compression.

\section{Related Work}
\label{Sec:RelatedWork}
Model compression technologies can be roughly classified as four categories: quantization \cite{rastegari2016quantization}, tensor factorization \cite{gong2014compressing,jaderberg2014factorization,kim2015factorization,lebedev2014factorization,xue2013factorization}, compact network design \cite{howard2017mobilenet,zhang2017shufflenet,figurnov2016perforatedcnns}, and channel pruning \cite{heeccv2018,hu2016nettrim,li2017filter,yu2018nisp,lemaire2019structured,zhao2019variational,ding2019centripetal}. The \textbf{quantization} approaches directly represent float points by using smaller number of bits. For example, the work in \cite{rastegari2016quantization} quantizes the network parameters into +1/-1 to accelerate computation. The \textbf{tensor factorization} methods decompose the weights into several low-rank matrices. In \cite{jaderberg2014factorization}, one 3$\times$3 filter is decomposed into one 1$\times$3 and one 3$\times$1 matrix, while in \cite{zhang2015factorization}, each layer is decomposed into a 3$\times$3 and a 1$\times$1 matrix. The \textbf{compact network design} approaches accelerate deep models by designing an efficient network architecture. The work in \cite{howard2017mobilenet} introduced the depth-wise convolution operation to accelerate the inference speed. In \cite{zhang2017shufflenet}, group-wise convolution is used to reduce computation of conventional convolution operations.

The \textbf{channel pruning} methods target at pruning a pre-defined number of channels and the corresponding weights related to these channels. For example, in \cite{liu2017learning}, the authors used the batch normalization layer to scale each channel and prune the channels with small scaling factors. In the width-multiplier method \cite{howard2017mobilenet}, the network is shrank uniformly and trained from scratch. More recently, the work in \cite{zhuang2018discrimination} proposed to select the channels according to the discriminative power of each channel, while the work in \cite{lin2019towards} used adversarial training to prune the channels automatically. The other channel pruning methods can be roughly categorized as two groups: loss minimization methods \cite{molchanov2017taylor,molchanov2019importance} and layer-by-layer approaches \cite{luoiccv2017,he2017channel}. On one hand, for loss minimization methods, Molchanov et al. \cite{molchanov2017taylor} aims at minimizing the loss change by iteratively removing the least important channels. When compared with the loss minimization methods, we do not need to frequently perform the time-consuming fine-tuning process, which makes our approach more efficient. In addition, in order to save the computational costs, the loss minimization approaches can only fine-tune the network under compression for a limitted number of iterations between two pruning stages, which cannot guarantee convergence. This leads to inaccurate evaluation of channel importance in the next pruning stage, which will degrade the channel selection performance. Our method selects the channels based on well-learned weights, which makes the channel selection in our approach more accurate. On the other hand, for the layer-by-layer approaches, He et al. \cite{he2017channel} proposed a channel pruning method to select the channels by solving a LASSO optimization problem and update the corresponding weights by solving a least square optimization problem in a layer-by-layer fashion. Our work also uses a similar approach to select the channels and update the weights, which makes our approach fast. However, our approach additionally considers the final loss in the channel pruning process and solves the \textit{next-layer feature removal} problem by paying more attention to the important features, which are important factors that influence the channel selection process but not considered in \cite{he2017channel}.

\section{Channel Pruning Guided by Classification Loss and Feature Importance}
\label{Sec:CPLI}
In this section, we firstly introduce the overview of our CPLI approach. Then, we present our method in the case of pruning a single layer in details. Finally, we present the pseudo code of the channel pruning process of the proposed approach.

\subsection{Overview of Our CPLI Framework}
\label{Sec:Overview}
Given a pre-trained model, our goal is to compress this model to achieve the highest accuracy for a given compression ratio. The pre-trained model is called the uncompressed model and the model after model compression 
is called the compressed model. 

Firstly, we extract the features of the uncompressed model based on the training dataset. Then, we prune the given uncompressed model in a layer-by-layer fashion from shallow layers (closer to the image) to deep layers (closer to the output) by using the objective function introduced in Eq. (\ref{eqn:objfinal}). After the channel pruning process, we perform the fine-tuning process on the compressed model to recover from the accuracy drop.

\subsection{CPLI in Each Layer}
\label{Sec:CPLI in layer}
For better presentation, we introduce our CPLI approach in each layer in this section. We firstly introduce the objective function of our CPLI method. Then, we will explain each part of our objective function in details.

\paragraph{(a) Formulation}
\subsubsection{}
Suppose we have a network with $L$ layers. When we have an input image $\mathcal{I}$, we can obtain the features at each layer in the network. Let us denote $\mathbf{X}^{0, (l)} \in \mathbb{R}^{c^{(l)}_{in} \times h^{(l)}_{in} \times w^{(l)}_{in}}$ as the input feature at the $l$-th layer of the uncompressed model, where $c^{(l)}_{in}$ is the number of input channels at this layer. $h^{(l)}_{in}$ and $w^{(l)}_{in}$ are the height and the width of the input feature $\mathbf{X}^{0, (l)}$, respectively. Denote $\mathbf{Y}^{0, (l)} \in \mathbb{R}^{c^{(l)}_{out} \times h^{(l)}_{out} \times w^{(l)}_{out}}$ as the output feature at the $l$-th layer, where $c^{(l)}_{out}$ is the number of output channels at this layer. $h^{(l)}_{out}$ and $w^{(l)}_{out}$ are the height and the width of the output feature $\mathbf{Y}^{0, (l)}$, respectively. Denote $\mathbf{W}^{0, (l)} \in \mathbb{R}^{c^{(l)}_{out} \times c^{(l)}_{in} \times h^{(l)}_{k} \times w^{(l)}_{k}}$ as the weights at the $l$-th layer of the uncompressed model. The $l$-th convolutional layer connects the features $\mathbf{X}^{0,(l)}$ and $\mathbf{Y}^{0,(l)}$. The output feature $\mathbf{Y}^{0,(l)}$ can be calculated as:
\begin{equation}
\label{eqn:conv}
\mathbf{Y}^{0,(l)}_{i,:,:} = \sum_{j=1}^{c^{(l)}_{in}}\beta^{0,(l)}_{j}\mathbf{X}^{0,(l)}_{j,:,:} * \mathbf{W}^{0,(l)}_{i,j,:,:},
\end{equation}
where $\mathbf{X}^{0, (l)}_{j,:,:}$ is the $j$-th channel of the input feature $\mathbf{X}^{0, (l)}$. $\mathbf{Y}^{0, (l)}_{i,:,:}$ is the $i$-th channel of the output feature $\mathbf{Y}^{0, (l)}$. $\mathbf{W}^{0, (l)}_{i,j,:,:}$ is the $j$-th channel of the $i$-th convolutional filter from $\mathbf{W}^{0, (l)}$. $*$ is the convolution operation. $\bm{\beta}^{0,(l)} \in \{0, 1\}^{c^{(l)}_{in}}$ is the vector containing a set of binary channel selection indicators $\beta^{0,(l)}_j, j= 1, \ldots, c^{(l)}_{in}$. For the uncompressed model, $\beta^{0, (l)}_j = 1$ for $j= 1, \ldots, c^{(l)}_{in}$, namely, all the channels are preserved. We omit the activation function and the bias term for better representation.

After pruning the previous layers, the $l$-th layer will have an input feature $\mathbf{X}^{(l)}$. Then we prune some channels of the input feature in the $l$-th layer and the compressed model will have an output feature $\mathbf{Y}^{(l)}$ at the $l$-th layer. Denote $\mathbf{W}^{(l)} \in \mathbb{R}^{c^{(l)}_{out} \times c^{(l)}_{in} \times h^{(l)}_{k} \times w^{(l)}_{k}}$ as the convolutional filters at the $l$-th layer in the compressed model, the output feature $\mathbf{Y}^{(l)}$ can be calculated as:
\begin{equation}
\label{eqn:conv2}
\mathbf{Y}^{(l)}_{i,:,:} = \sum_{j=1}^{c^{(l)}_{in}}\beta^{(l)}_{j}\mathbf{X}^{(l)}_{j,:,:} * \mathbf{W}^{(l)}_{i,j,:,:},
\end{equation}
where $\mathbf{X}^{(l)}_{j,:,:}$ is the $j$-th channel of the input feature $\mathbf{X}^{(l)}$. $\mathbf{Y}^{(l)}_{i,:,:}$ is the $i$-th channel of the output feature $\mathbf{Y}^{(l)}$.  $\mathbf{W}^{(l)}_{i,j,:,:}$ is the $j$-th channel of the $i$-th convolutional filter from $\mathbf{W}^{(l)}$. The other notations are the same as those in Eq. (\ref{eqn:conv}). $\bm{\beta}^{(l)} \in \{0, 1\}^{c^{(l)}_{in}}$ is the vector containing a set of binary channel selection indicators $\beta^{(l)}_j, j= 1, \ldots, c^{(l)}_{in}$. If $\beta^{(l)}_{j}=0$, the $j$-th channel of the input feature at layer $l$ will be pruned.

The layer-by-layer channel pruning approach in the work \cite{he2017channel} prunes the channels by solving the following optimization problem:
\begin{eqnarray}
\label{eqn:cp}
\begin{aligned}
&\argmin_{\bm{\beta}^{(l)},\mathbf{W}^{(l)}} \|\mathbf{Y}^{0,(l)}-\mathbf{Y}^{(l)}\|_{F}^{2}, \\
=&\argmin_{\bm{\beta}^{(l)},\mathbf{W}^{(l)}} \sum_{i=1}^{c^{(l)}_{out}} \sum_{m=1}^{M^{(l)}} \left[y_{i,m}^{0,(l)}-y^{(l)}_{i,m}\right]^{2}, \\
&\!\!\!\text{subject to } \|\bm{\beta}^{(l)}\|_{0} \leq B^{(l)},
\end{aligned}
\end{eqnarray}
where $y_{i,m}^{0,(l)}$ and $y_{i,m}^{(l)}$ are the output features of the $i$-th channel at location $m$ for the uncompressed model and the compressed model, respectively. $y_{i,m}^{0,(l)}$ is an element in $\mathbf{Y}^{0,(l)}$ and $y_{i,m}^{(l)}$ is an element in $\mathbf{Y}^{(l)}$. $M^{(l)}$ is the length of the vectorized output feature map at the $l$-th layer. $B^{(l)}$ is the pre-defined number of remained channels for the $l$-th layer. $\|\cdot\|_{0}$ is $l_{0}$ norm and $\|\cdot\|_{F}$ is the Frobenius norm.

The objective function in Eq. (\ref{eqn:cp}) only minimizes the reconstruction error between $y_{i,m}^{0,(l)}$ and $y^{(l)}_{i,m}$. It does not take the final classification loss or the feature importance into account. Therefore, we propose our objective function by additionally considering the final classification loss and feature importance.

\subsubsection{Notation change.}
Since we focus on how to prune the $l$-th layer of the network in this section, we drop the layer index $l$ thereafter except in Algorithm \ref{alg:framework} for better presentation.

\subsubsection{Classification loss.}
Let us denote $\mathcal{C}(\mathbf{Y}, g;\mathcal{W})$ as the classification loss function of the compressed network when the output feature at the $l$-th layer is $\mathbf{Y}$. The classification loss function of the compressed model can be defined as follows:
\begin{eqnarray}
\begin{aligned}
\label{eqn:cost}
&\mathcal{C} = \mathcal{L}_{c} \left[\mathcal{N}(\mathbf{Y};\mathcal{W}),g \right], \\
\end{aligned}
\end{eqnarray}
where $\mathcal{L}_{c}$ is the cross-entropy loss function. $g$ is the ground truth label for the image $\mathcal{I}$. When pruning the $l$-th layer, $\mathcal{N}$ and $\mathcal{W}$ are the sub-network and its parameters from the ($l+1$)-th layer to the $L$-th layer in the compressed model, respectively.

\subsubsection{Our overall objective function.}
Denote $y_{i,m}$ for $i=1, \ldots, c_{out}$, $m=1, \ldots, M$ as an element in $\mathbf{Y}$. In our CPLI approach, we prune the channels by considering the final classification loss and the feature importance. To this end, we propose the following objective function for each layer:
\begin{eqnarray}
\label{eqn:objfinal}
\begin{aligned}
&\argmin_{\bm{\beta},\mathbf{W}} \sum_{i=1}^{c_{out}} \sum_{m=1}^{M} \left[\frac{\partial \mathcal{C}}{\partial y_{i,m}} \cdot (y_{i,m}^{0}-\gamma y_{i,m}^{*} \cdot y_{i,m})\right]^{2}, \\
&\text{subject to } \|\bm{\beta}\|_{0} \leq B,
\end{aligned}
\end{eqnarray}
where $\frac{\partial \mathcal{C}}{\partial y_{i,m}}$ is the partial derivative of the classification loss with respect to $y_{i,m}$, and the loss function $\mathcal{C}$ is defined in Eq.~(\ref{eqn:cost}). $y^{*}_{i,m}$ is the $i$-th channel at location $m$ from the output features in the compressed model after the pruning previous layers. $\gamma$ is a constant, which is empirically set as 1.
The term $\frac{\partial \mathcal{C}}{\partial y_{i,m}}$ corresponds to the guidance from the classification loss and the term $\gamma y_{i,m}^{*}$ corresponds to the guidance from the feature importance. We will introduce the motivation of these two terms in the following section.

\paragraph{(b) Guidance from the Classification Loss}
\subsubsection{Channel pruning with guidance from the classification loss.}
If we only consider the classification loss without considering the feature importance, the objective function in Eq. (\ref{eqn:objfinal}) is redefined as follows:
\begin{eqnarray}
\label{eqn:objcl}
\begin{aligned}
&\argmin_{\bm{\beta},\mathbf{W}} \sum_{i=1}^{c_{out}} \sum_{m=1}^{M} \left[\frac{\partial \mathcal{C}}{\partial y_{i,m}}\cdot(y_{i,m}^{0}-y_{i,m})\right]^{2}, \\
&\text{subject to } \|\bm{\beta}\|_{0} \leq B,
\end{aligned}
\end{eqnarray}
where $\frac{\partial \mathcal{C}}{\partial y_{i,m}}$ is the partial derivative of the classification loss with respect to $y_{i,m}$. $y_{i,m}^{0}$ and $y_{i,m}$ are the output features from the $i$-th channel at location $m$ for the uncompressed model and the compressed model, respectively. $M$ is the length of the vectorized output feature map. $B$ is the pre-defined number of remained channels for this layer. The other notations are the same as those in Eq. (\ref{eqn:objfinal}).

\subsubsection{Analysis.}
The term  $(y_{i,m}^{0}-y_{i,m})$ in Eq. (\ref{eqn:objcl}) can be considered as the reconstruction error, which is the error by using $y_{i,m}$ to reconstruct $y_{i,m}^{0}$. 
At the spatial location $m$, the term $\frac{\partial \mathcal{C}}{\partial y_{i,m}}$ can be treated as the ``importance'' of the $i$-th channel, which indicates the impact by changing $y_{i,m}$ on the classification loss $\mathcal{C}$. For the same reconstruction error, a large ``importance'' value indicates that even a small change of this channel will cause a large change of the classification loss. In this situation, this feature is important and we should pay more attention to how to minimize the reconstruction error of this channel. As another extreme, if $\frac{\partial \mathcal{C}}{\partial y_{i,m}}=0$, the $i$-th channel is not important and the reconstruction error does not contribute to the objective function for channel pruning.

When compared with the objective function in the work \cite{he2017channel}, we additionally consider the term $\frac{\partial \mathcal{C}}{\partial y_{i,m}}$, which can weight each channel by considering the impact of this channel on the final classification loss. In this way, our approach can be guided by the classification loss.

\paragraph{(c) Guidance from the Feature Importance}
\subsubsection{The \emph{next-layer feature removal} problem.} 
At the current pruning stage, we prune the input features at layer $l$, in which the output feature $y_{i,m}$ is used in the objective function to reconstruct $y_{i,m}^{0}$. In the next pruning stage, we will prune layer $l+1$, in which the feature $y_{i,m}$ will be treated as the input features and may be pruned. This is not taken into consideration in the previous section. If the $i$-th channel of the output features is removed when pruning the next layer $l+1$ at the next pruning stage, then $y_{i,m}=0$. In this case, reconstruction of $y_{i,m}$ is unnecessary and may lead to inaccurate channel selection. This problems is called as the \textit{next-layer feature removal} problem.

It is worth mentioning that the \emph{next-layer feature removal} problem often occurs when the compression ratio is large. For example, more than half of channels in each layer in a ResNet-50 model under 2.25$\times$ compression ratio is removed, which indicates that the probability of the \emph{next-layer feature removal} problem occurs is high. 

\subsubsection{Channel pruning with guidance from feature importance.}
To handle the \textit{next-layer feature removal} problem, we propose to introduce the term $\gamma y_{i,m}^{*}$ into the objective function in Eq. (\ref{eqn:objfinal}). If we only consider the feature importance without considering the classification loss, the objective function can be written as follows:
\begin{eqnarray}
\label{eqn:objfi}
\begin{aligned}
&\argmin_{\bm{\beta},\mathbf{W}} \sum_{i=1}^{c_{out}} \sum_{m=1}^{M} (y_{i,m}^{0}-\gamma y_{i,m}^{*} \cdot y_{i,m})^{2}, \\
&\text{subject to } \|\bm{\beta}\|_{0} \leq B.
\end{aligned}
\end{eqnarray}

This objective function can be equivalently rewritten as follows:
\begin{eqnarray}
\label{eqn:objfi2}
\begin{aligned}
&\argmin_{\bm{\beta},\mathbf{W}} \sum_{i=1}^{c_{out}} \sum_{m=1}^{M} \left[\gamma y_{i,m}^{*} \cdot (y_{i,m}^{0} - y_{i,m}) + \right.\\
&\qquad \qquad \qquad \quad \left.(1-\gamma y_{i,m}^{*}) \cdot (y_{i,m}^{0} - 0) \right]^{2}, \\
&\text{subject to } \|\bm{\beta}\|_{0} \leq B,
\end{aligned}
\end{eqnarray}
where $y^{*}_{i,m}$ is the output features of the $i$-th channel at location $m$ after pruning the previous layers and $\gamma$ is a constant, which is empirically set as 1 in our experiments. The other notations are the same as those in Eq. (\ref{eqn:objcl}).

At the spatial location $m$, we use $\gamma y^{*}_{i,m}$ as the guidance of whether the $i$-th channel will be removed or not in the next pruning stage.
When $\gamma y_{i,m}^{*}=0$ for most of the spatial locations $m$, it is more likely that the $i$-th channel will be removed at the next pruning stage. In this case, we will use the reconstruction error of ($y_{i,m}^{0}-0$). On the other hand, if the $i$-th channel is not pruned at the next stage, we will use the reconstruction error of ($y_{i,m}^{0}-y_{i,m}$). Since the removal of the channels at the next pruning stage is determined by many factors, including the spatial location $m$, the input image, and the learned parameters, it is hard to predict whether this channel will be removed at the next pruning stage. In this work, we empirically use $\gamma y^{*}_{i,m}$ as the guidance, which is based on the following two observations. 1) In the ideal case, if we have $\gamma y^{*}_{i,m}=0$ for all positions $m$ in the $i$-th channel, the $i$-th channel can be readily removed at the next pruning stage. 2) The magnitude of $\gamma y^{*}_{i,m}$ is also used as the guidance for preserving/pruning the neuron in \cite{han2015learning}. Although the aforementioned ideal case will rarely happen in real applications, our channel pruning method using $\gamma y^{*}_{i,m}$ as the guidance effectively avoids minimizing the reconstruction error from uninformative features that will be removed at the next pruning stage and thus can achieve reasonable results.

\paragraph{(d) Solution}
\subsubsection{Notation change.} 
For better presentation, thereafter, we omit the summation over spatial locations $M$ and the index for location $m$. The formulation for multiple locations can be readily obtained.

\subsubsection{Solution.}
It is a NP-hard problem to solve the objective function in Eq. (\ref{eqn:objfinal}). Therefore, we relax the $l_{0}$ regularization to $l_{1}$ regularization and arrive at the following objective function:
\begin{eqnarray}
\label{eqn:objlasso}
\begin{aligned}
&\argmin_{\bm{\beta},\mathbf{W}} \sum_{i=1}^{c_{out}} \left[\frac{\partial \mathcal{C}}{\partial y_{i}} \cdot (y_{i}^{0}- \gamma y_{i}^{*} \cdot y_{i})\right]^{2} + \lambda \|\bm{\beta}\|_{1}, \\
&\text{subject to } \|\bm{\beta}\|_{0} \leq B,
\end{aligned}
\end{eqnarray}
where $\lambda$ is the coefficient to balance different terms. Following \cite{he2017channel}, we solve the optimization problem in Eq. (\ref{eqn:objlasso}) by using two steps. First, we fix $\mathbf{W}$ and solve a LASSO optimization problem:
\begin{eqnarray}
\label{eqn:lasso}
\begin{aligned}
&\argmin_{\bm{\beta}}\sum_{i=1}^{c_{out}} \left[\frac{\partial \mathcal{C}}{\partial y_{i}} \cdot (y_{i}^{0}- \gamma y_{i}^{*} \cdot y_{i})\right]^{2} + \lambda \|\bm{\beta}\|_{1}, \\
&\text{subject to } \|\bm{\beta}\|_{0} \leq B.
\end{aligned}
\end{eqnarray}

We gradually increase $\lambda$ until the constraint $\|\bm{\beta}\|_{0}\leq B$ is satisfied. After the channel selection process is finished, we treat each selected channel equally important because the remaining channels have large impact on the output of the classification loss (\textit{i.e.}, the channels with small impact are already pruned). Therefore, we consider $\gamma y_{i}^*=1$, drop the term $\frac{\partial \mathcal{C}}{\partial y_{i}}$, and minimize the reconstruction error by solving a least square optimization problem with fixed $\bm{\beta}$: 
\begin{equation}
\label{eqn:ls}
\argmin_{\mathbf{W}}\sum_{i=1}^{c_{out}}(y_{i}^{0}-y_{i})^{2}.
\end{equation}
In this way, we minimize the objective function for each layer. We iteratively perform the pruning process in a layer-by-layer fashion to compress the pre-trained model and obtain the compressed model before the fine-tuning process.

\subsection{Pseudo Code}
Algorithm \ref{alg:framework} presents the pseudo code of our CPLI approach for pruning a pre-trained model. Given a pre-trained model, we can use Algorithm \ref{alg:framework} to prune this model, and obtain the compressed model $\mathrm{M}_c$ before the fine-tuning process. Again, for better representation, we omit the summation over spatial locations $M$ and the index for location $m$ when introducing our algorithm. In practice, we do not prune the first layer of the uncompressed model because the input features of the first layer are raw images. Then, we fine-tune the compressed model to recover from the accuracy drop.
{\SetAlgoNoLine
\begin{algorithm}[t]
  \caption{Our CPLI approach for pruning the pre-trained model.} 
  \label{alg:framework}  
    \KwIn{Pre-trained model $\mathrm{M}_{u}$, $\mathrm{M}_{u} = \{\mathbf{W}^{0,(1)}, \mathbf{W}^{0, (2)},\ldots,\mathbf{W}^{0,(L)}, \bm{\Theta}\}$, where $\mathbf{W}^{0,(l)}$ for $l \in L$ is the parameters for the $l$-th layer and $\bm{\Theta}$ is the parameters for other layers (\textit{e.g.}, the fully connected layers) that will not be pruned. 
    }
    \KwOut{Compressed model $\mathrm{M}_{c}$, which is then used for the fine-tuning process.}
    Extract the output features in the pre-trained model for each layer $y_i^{0,(1)},\ldots,y_i^{0,(L)}$ for all channels.
    
    Set $\mathrm{M}_c=\mathrm{M}_u= \{\mathbf{W}^{0,(1)}, \mathbf{W}^{0,(2)},\ldots,\mathbf{W}^{0,(L)}, \bm{\Theta}\}$.
    
    \For {l = 2 $to$ L }
    {
        Based on the current compressed model $\mathrm{M}_c$, use forward-propagation to calculate ${y_i^{*}}^{(l)}$ in Eq. (\ref{eqn:lasso}), where the superscript $\cdot^{(l)}$ denotes the $l$-th layer.
        
        Use back-propagation to calculate $\frac{\partial \mathcal{C}}{\partial y_i^{(l)}}$ in Eq. (\ref{eqn:lasso}), where $y_i^{(l)}$ is a feature of the current compressed model at layer $l$ and channel $i$. 
        
        Solve the LASSO optimization problem in Eq. (\ref{eqn:lasso}), and obtain the channel selection vector $\bm{\beta}^{(l)}$ for the $l$-th layer.
        
        Solve the the least square optimization problem in Eq. (\ref{eqn:ls}) with fixed $\bm{\beta}^{(l)}$, and obtain the adjusted weights $\Tilde{\mathbf{W}}^{(l)}$ for the $l$-th layer.
        
        Prune the filters in the ($l-1$)-th layer by removing the $k$-th filter in $\Tilde{\mathbf{W}}^{(l-1)}$ where $k$ are the indices for all $\beta^{(l)}_{k}=0$ in $\bm{\beta}^{(l)}$ and obtain $\hat{\mathbf{W}}^{(l-1)}$.
        
        Prune the channels by setting $\mathrm{M}_c =\{\hat{\mathbf{W}}^{(1)}, \hat{\mathbf{W}}^{(2)}, \ldots, \hat{\mathbf{W}}^{(l-1)},$  $\Tilde{\mathbf{W}}^{(l)}, \mathbf{W}^{0,(l+1)},\ldots, \mathbf{W}^{0,(L)}, \bm{\Theta}\}$
    }
    Obtain the compressed model $\mathrm{M}_c$ before the fine-tuning process.
\end{algorithm} 
}

\section{Experiments}
\label{sec:experiments}
In this section, we first compare our CPLI approach with several state-of-the-art channel pruning methods on two benchmark datasets: CIFAR-10 \cite{cifar10} and ImageNet \cite{imagenet} for the image classification task.
We further conduct the experiments to prune 3D convolutional network for the action recognition task on the UCF-101 dataset \cite{soomro2012ucf101} to demonstrate the generalization ability of our method. 
We finally investigate each component of our method in details.

The compression ratio (CR) refers to the ratio of the floating point operations (FLOPs) from the uncompressed model over that from the compressed model, which is a commonly used criterion for computational complexity measurement. Since the accuracies of the pre-trained model vary a lot for different baseline methods, we follow the work in \cite{zhuang2018discrimination} to report the accuracy drop after the fine-tuning process.

\begin{table*}[t]
\centering
\caption{Comparison of different channel pruning methods for compressing VGGNet, ResNet-56, and MobileNet-V2 on the CIFAR-10 dataset. For reference, in our implementation, the accuracies of the uncompressed VGGNet, ResNet-56, and MobileNet-V2 models are 93.99\%, 93.74\%, and 95.02\%, respectively. ``+'' denotes that the accuracy increases, while ``-'' denotes that the accuracy decreases after channel pruning. We directly quote the results of the existing works from \cite{zhuang2018discrimination}.}
\begin{tabular}{c|c||cccccc}
\toprule[1pt]
\multicolumn{2}{c||}{Model} & ThiNet & CP & Slimming & WM & DCP & \textbf{Ours}\\
\midrule[1pt]
\multirow{2}{*}{VGGNet} & CR & 2.00$\times$ & 2.00$\times$ & 2.04$\times$ & 2.00$\times$ & 2.86$\times$ & \textbf{2.86$\times$} \\
 & Top-1 Acc. drop (\%) & -0.14 & -0.32 & -0.19 & -0.38 & +0.58 & \textbf{+0.96} \\
\hline
\multirow{2}{*}{ResNet-56} & CR & 1.99$\times$ & 2$\times$ & - & 1.99$\times$ & 1.99$\times$ & \textbf{1.99$\times$} \\
 & Top-1 Acc. drop (\%) & -0.82 & -1.0 & - & -0.56 & -0.31 & \textbf{+0.07} \\
\hline
\multirow{2}{*}{MobileNet-V2} & CR & - & - & - & 1.36$\times$ & 1.36$\times$ & \textbf{1.36$\times$} \\
 & Top-1 Acc. drop (\%) & - & - & - & -0.45 & +0.22 & \textbf{+0.35} \\
\bottomrule[1pt]
\end{tabular}
\label{tab:VGGResNet}
\end{table*}

\begin{table*}[t]
\centering
\caption{Comparison of Top-5 accuracies when compressing ResNet-50 and Top-1 accuracies when compressing MobileNet-V2 by using different channel pruning methods on the ImageNet dataset. For reference, in our implementation, the Top-5 accuracy of the uncompressed ResNet-50 model and Top-1 accuracy of the uncompressed MobileNet-V2 model on the ImageNet dataset are 92.87\% and 72.19\%, respectively. For the existing works, we directly quote the results from \cite{zhuang2018discrimination}.}
\begin{tabular}{c|c||cccccc}
\toprule[1pt]
\multicolumn{2}{c||}{Model} & ThiNet & CP & WM & DCP & GAL & \textbf{Ours}\\
\midrule[1pt]
\multirow{2}{*}{ResNet-50} & CR & 2.25$\times$ & 2$\times$ & 2.25$\times$ & 2.25$\times$ & 2.22$\times$ & \textbf{2.25$\times$} \\
 & Top-5 Acc. drop (\%) & -1.12 & -1.40 & -1.62 & -0.61 & -2.05 & \textbf{-0.38} \\
\hline
\multirow{2}{*}{MobileNet-V2} & CR & 1.81$\times$ & - & 1.81$\times$ & 1.81$\times$ & - & \textbf{1.81$\times$} \\
 & Top-1 Acc. drop (\%) & -6.36 & - & -6.40 & -5.89 & - & \textbf{-4.84} \\
\bottomrule[1pt]
\end{tabular}
\label{tab:ImageNet}
\end{table*}

\subsection{Results on CIFAR-10}
\label{sec:cifar10}
We take three popular models VGGNet \cite{simonyan2014very}, ResNet-56 \cite{he2016deep}, and MobileNet-V2 \cite{sandler2018mobilenetv2} on the CIFAR-10 dataset to demonstrate the effectiveness of the proposed approach. The CIFAR-10 dataset \cite{cifar10} consists of 50k training samples and 10k testing images from 10 classes. Based on the pre-trained models, we apply our proposed CPLI method to prune the channels and fine-tune the compressed model. In the pruning process, we randomly choose 10 locations from each feature map instead of using all the spatial locations to accelerate the pruning process. At the fine-tuning stage, similar to \cite{zhuang2018discrimination}, we use SGD with nesterov for optimization. The momentum, the weight decay, and the mini-batch size are set to 0.9 and 0.0001, and 256, respectively. The initial learning rate is set to 0.1 and step learning rate decay is used.

In Table \ref{tab:VGGResNet}, we compare our method with several state-of-the-art methods including ThiNet \cite{luoiccv2017}, Channel Pruning (CP) \cite{he2017channel}, Slimming \cite{liu2017learning}, Width-multiplier (WM) \cite{howard2017mobilenet}, and DCP \cite{zhuang2018discrimination} for compressing VGGNet, ResNet-56, and MobileNet-V2 on the CIFAR-10 dataset. 

From the results in Table \ref{tab:VGGResNet}, our CPLI approach consistently outperforms other baseline methods, which demonstrates the effectiveness of our method on small-scale datasets.
It is also worth mentioning that our CPLI approach outperforms the pre-trained VGGNet, ResNet-56, and MobileNet-V2 by 0.96\%, 0.07\%, and 0.35\%, respectively. Similar results are also reported in the DCP work \cite{zhuang2018discrimination}. We hypothesize that the overfitting problem on small-scale datasets like CIFAR-10 can be partially solved by pruning redundant channels.

\subsection{Results on ImageNet}
\label{sec:imagenet}
To evaluate the effectiveness of our CPLI approach on large-scale datasets, we further conduct the experiments to prune ResNet-50 \cite{he2016deep} and MobileNet-V2 \cite{sandler2018mobilenetv2} on the ILSVRC-12 dataset \cite{imagenet}. The ILSVRC-12 dataset is a large-scale dataset, which contains 1.28 million training images and 50k testing images from 1000 categories. 
Due to the shortcut design in the residual block, we cannot directly prune the first layer of the residual block. Therefore, we follow the method in \cite{he2017channel} to prune the multi-branch networks like ResNet. 
For ResNet-50, we follow the setting in \cite{lin2019towards} to fine-tune the pruned model with hint \cite{romero2014fitnets} from the last layer. The initial learning rate is set to 0.01 and the batch size is set to 128. The other settings are the same as those on the CIFAR-10 dataset. For MobileNet-V2, we use the same setting as that for ResNet-50 except that the batch size is set to 256.

Similar to the experiments on CIFAR-10, we compare our proposed approach with ThiNet \cite{luoiccv2017}, Channel Pruning (CP) \cite{he2017channel}, Width-multiplier (WM) \cite{howard2017mobilenet}, GAL \cite{lin2019towards} and DCP \cite{zhuang2018discrimination} in Table \ref{tab:ImageNet}. 
Consistent with the experiments on the CIFAR-10 dataset, for ResNet-50, our CPLI approach outperforms DCP and GAL by 0.23\% and \textbf{1.67\%}, respectively. 

Since the work in \cite{zhuang2018discrimination} only reports the Top-1 accuracy of the uncompressed model for MobileNet-V2, we follow the setting in \cite{zhuang2018discrimination} and report the Top-1 accuracy drop after the model compression process. Again, our proposed approach surpasses the baseline methods ThiNet, WM, and DCP by \textbf{1.52\%}, \textbf{1.56\%}, and \textbf{1.05\%}, respectively, which are \textbf{significant improvements} on the ImageNet dataset. The results on the ImageNet dataset clearly demonstrate that it is beneficial to prune channel by using our approach on the large-scale datasets.

\subsection{Results on UCF-101}
\label{sec:ucf101}
In order to demonstrate the generalization ability of our CPLI approach, we further compress the C3D model \cite{tran2015learning} on the UCF-101 dataset \cite{soomro2012ucf101} for the action recognition task. The UCF-101 dataset consists of 13320 videos with the resolution of 320 $\times$ 240. We uniformally extract a set of frames from the videos at the rate of 25 fps. In the fine-tuning process, we resize the frames into 128 $\times$ 171 and randomly crop the frames to the resolution of 112 $\times$ 112. For testing, all frames are center cropped to the resolution of 112 $\times$ 112. We also split the frames of each video into several non-overlapped clips, where each clip contains 16 frames. These clips are used as the inputs of the C3D network. 

Following \cite{zhang2018three}, we compare our method with Taylor Pruning (TP) \cite{molchanov2017taylor}, Filter Pruning (FP) \cite{li2017filter}, and Regularization-based pruning (RBP) \cite{zhang2018three} in terms of the clip-level accuracy drop. 
The work in FP selects the channels based on the magnitude of the features. In RBP, the authors add the regularization terms to the loss function and remove the channels based on the $l_{1}$ norm of the weights. The results are shown in Table \ref{tab:C3D}. Again, our CPLI method outperforms the baseline methods TP, FP, and RBP, which demonstrates the generalization ability of our proposed method for compressing the C3D model.
\begin{table}[t]
\centering
\caption{Clip-level accuracy drops after compressing C3D on UCF-101 by using different channel pruning approaches. The clip-level accuracy of the pre-trained model is 79.93\%. We directly quote the results from \cite{zhang2018three}.}
\resizebox{\textwidth}{!}{ 
\begin{tabular}{c|c||cccccc}
\toprule[1pt]
\multicolumn{2}{c||}{Model} & TP & FP & RBP & \textbf{Ours} \\
\midrule[1pt]
\multirow{2}{*}{C3D} & CR & 2$\times$ & 2$\times$ & 2$\times$ & \textbf{2$\times$} \\
 & Clip-level Acc. drop (\%) & -11.50 & -4.92 & -3.56 & \textbf{-2.18} \\
\bottomrule[1pt]
\end{tabular}}
\label{tab:C3D}
\end{table}

\subsection{Discussion}
The results on CIFAR-10, ImageNet and UCF-101 clearly demonstrate the effectiveness of our CPLI approach for different datasets, network structures, and tasks. Our method performs better than the loss minimization methods, such as TP, because the channel selection process is based on the weights learned by solving the least square optimization problem. It converges better than the alternative approaches where the weights are learned by several rounds of fine-tuning. Compared with the layer-by-layer methods \cite{he2017channel,luoiccv2017}, the performance improvement comes from the consideration of the final loss and the feature importance.

\subsection{Ablation Study}
\label{sec:ablation}
In this section, we take pruning VGGNet under 2.86$\times$ compression ratio on the CIFAR-10 dataset as an example to investigate each component in our CPLI approach.

\begin{table}[t]
\centering
\caption{Accuracy drops after pruning VGGNet under 2.86$\times$ compression ratio on CIFAR-10 without considering the final loss (FL) and the feature importance (FI). For reference, in our implementation, the accuracy of the uncompressed VGGNet model is 93.99\%. ``+'' denotes that the accuracy increases.}
\begin{tabular}{c||c}
\toprule[1pt]
Methods & Acc. drop (\%) \\
\midrule[1pt]
CPLI (Baseline) & +0.96 \\
CPLI w/o FL & +0.04 \\
CPLI w/o FI & +0.47 \\
\bottomrule[1pt]
\end{tabular}
\label{tab:ablation}
\end{table}

\subsubsection{Effect of the classification loss.} 
To investigate the effectiveness after taking the classification loss into account in the channel pruning process, we prune the channels without considering the final loss by setting the term $\frac{\partial \mathcal{C}}{\partial y_{i}}$ in Eq. (\ref{eqn:objlasso}) to 1, which is referred as CPLI w/o FL in Table \ref{tab:ablation}. In this case, we only address the \emph{next-layer feature removal} problem during the channel pruning process but ignore the final loss. In Table \ref{tab:ablation}, the accuracy drops 0.92\% when compared with our proposed method, which shows that it is important to take the final loss into account in the channel pruning process.

\subsubsection{Effect of feature importance.} 
To investigate the effectiveness after considering the feature importance in the channel pruning process, we set the term $\gamma y_{i}^{*}$ in Eq. (\ref{eqn:objlasso}) as 1, which is denoted by CPLI w/o FI in Table \ref{tab:ablation}. In this case, we only consider the classification loss but ignore the \emph{next-layer feature removal} problem in the channel pruning process. In Table \ref{tab:ablation}, the accuracy drops 0.49\% when compared with the proposed method, which suggests that it is beneficial to address the \emph{next-layer feature removal} issue.

\subsubsection{Results using different number of sampled locations.}  
We conduct more experiments to investigate the performance when using different number of sampled locations. The results are shown in Table \ref{tab:samples}. From Table \ref{tab:samples}, we observe that the accuracies increase when the number of spatial locations increases from 1 to 10, but the accuracies are almost the same when the number of spatial locations increases from 10 to 20. The results indicate that it is sufficient to use 10 spatial locations in the pruning process. Therefore, we choose 10 spatial locations to prune the original model for the trade-off between accuracy and speed.
\begin{table}[t]
\centering
\caption{Comparison of accuracy drops after pruning VGGNet with 2.86$\times$ compression ratio on CIFAR-10 when using different number of sampled spatial locations. ``+'' denotes that the accuracy increases.}
\begin{tabular}{c||c}
\toprule[1pt]
Number of points & Acc. drop (\%) \\
\midrule[1pt]
1 & +0.12 \\
5 & +0.77 \\
10 & +0.96 \\
20 & +0.97 \\
\bottomrule[1pt]
\end{tabular}
\label{tab:samples}
\end{table}

\section{Conclusion}
In this work, we have proposed a new channel pruning approach called Channel Pruning guided by classification Loss and feature Importance (CPLI) to compress CNNs. Our method take the classification loss into account without frequently performing the fine-tuning process. We have also proposed an effective approach to address the \emph{next-layer feature removal} problem, in which the uninformative features can be ignored in the pruning process. Comprehensive experiments on three benchmark datasets clearly demonstrate the effectiveness of our newly proposed CPLI approach for model compression.

\bibliographystyle{aaai}
\bibliography{egbib}

\begin{thebibliography}{}

\bibitem[\protect\citeauthoryear{Ding \bgroup et al\mbox.\egroup
  }{2019}]{ding2019centripetal}
Ding, X.; Ding, G.; Guo, Y.; and Han, J.
\newblock 2019.
\newblock Centripetal sgd for pruning very deep convolutional networks with
  complicated structure.
\newblock In {\em CVPR}.

\bibitem[\protect\citeauthoryear{Figurnov \bgroup et al\mbox.\egroup
  }{2016}]{figurnov2016perforatedcnns}
Figurnov, M.; Ibraimova, A.; Vetrov, D.~P.; and Kohli, P.
\newblock 2016.
\newblock Perforatedcnns: Acceleration through elimination of redundant
  convolutions.
\newblock In {\em NeurIPS}.

\bibitem[\protect\citeauthoryear{Gong \bgroup et al\mbox.\egroup
  }{2014}]{gong2014compressing}
Gong, Y.; Liu, L.; Yang, M.; and Bourdev, L.
\newblock 2014.
\newblock Compressing deep convolutional networks using vector quantization.
\newblock {\em arXiv preprint arXiv:1412.6115}.

\bibitem[\protect\citeauthoryear{Han \bgroup et al\mbox.\egroup
  }{2015}]{han2015learning}
Han, S.; Pool, J.; Tran, J.; and Dally, W.
\newblock 2015.
\newblock Learning both weights and connections for efficient neural network.
\newblock In {\em NeurIPS}.

\bibitem[\protect\citeauthoryear{He \bgroup et al\mbox.\egroup
  }{2016}]{he2016deep}
He, K.; Zhang, X.; Ren, S.; and Sun, J.
\newblock 2016.
\newblock Deep residual learning for image recognition.
\newblock In {\em CVPR}.

\bibitem[\protect\citeauthoryear{He \bgroup et al\mbox.\egroup
  }{2018}]{heeccv2018}
He, Y.; Lin, J.; Liu, Z.; Wang, H.; Li, L.-J.; and Han, S.
\newblock 2018.
\newblock {AMC}: Automl for model compression and acceleration on mobile
  devices.
\newblock In {\em ECCV}.

\bibitem[\protect\citeauthoryear{He, Zhang, and Sun}{2017}]{he2017channel}
He, Y.; Zhang, X.; and Sun, J.
\newblock 2017.
\newblock Channel pruning for accelerating very deep neural networks.
\newblock In {\em ICCV}.

\bibitem[\protect\citeauthoryear{Howard \bgroup et al\mbox.\egroup
  }{2017}]{howard2017mobilenet}
Howard, A.~G.; Zhu, M.; Chen, B.; Kalenichenko, D.; Wang, W.; Weyand, T.;
  Andreetto, M.; and Adam, H.
\newblock 2017.
\newblock Mobilenets: Efficient convolutional neural networks for mobile vision
  applications.
\newblock {\em arXiv preprint arXiv:1704.04861}.

\bibitem[\protect\citeauthoryear{Hu \bgroup et al\mbox.\egroup
  }{2016}]{hu2016nettrim}
Hu, H.; Peng, R.; Tai, Y.-W.; and Tang, C.-K.
\newblock 2016.
\newblock Network trimming: A data-driven neuron pruning approach towards
  efficient deep architectures.
\newblock {\em arXiv preprint arXiv:1607.03250}.

\bibitem[\protect\citeauthoryear{Jaderberg, Vedaldi, and
  Zisserman}{2014}]{jaderberg2014factorization}
Jaderberg, M.; Vedaldi, A.; and Zisserman, A.
\newblock 2014.
\newblock Speeding up convolutional neural networks with low rank expansions.
\newblock In {\em BMVC}.

\bibitem[\protect\citeauthoryear{Kim \bgroup et al\mbox.\egroup
  }{2015}]{kim2015factorization}
Kim, Y.-D.; Park, E.; Yoo, S.; Choi, T.; Yang, L.; and Shin, D.
\newblock 2015.
\newblock Compression of deep convolutional neural networks for fast and low
  power mobile applications.
\newblock {\em arXiv preprint arXiv:1511.06530}.

\bibitem[\protect\citeauthoryear{Krizhevsky}{2009}]{cifar10}
Krizhevsky, A.
\newblock 2009.
\newblock Learning multiple layers of features from tiny images.
\newblock Technical report, Citeseer.

\bibitem[\protect\citeauthoryear{Lebedev \bgroup et al\mbox.\egroup
  }{2014}]{lebedev2014factorization}
Lebedev, V.; Ganin, Y.; Rakhuba, M.; Oseledets, I.; and Lempitsky, V.
\newblock 2014.
\newblock Speeding-up convolutional neural networks using fine-tuned
  cp-decomposition.
\newblock {\em arXiv preprint arXiv:1412.6553}.

\bibitem[\protect\citeauthoryear{LeCun, Denker, and
  Solla}{1990}]{lecun1990optimal}
LeCun, Y.; Denker, J.~S.; and Solla, S.~A.
\newblock 1990.
\newblock Optimal brain damage.
\newblock In {\em NeurIPS}.

\bibitem[\protect\citeauthoryear{Lemaire, Achkar, and
  Jodoin}{2019}]{lemaire2019structured}
Lemaire, C.; Achkar, A.; and Jodoin, P.-M.
\newblock 2019.
\newblock Structured pruning of neural networks with budget-aware
  regularization.
\newblock In {\em CVPR}.

\bibitem[\protect\citeauthoryear{Li \bgroup et al\mbox.\egroup
  }{2016}]{li2017filter}
Li, H.; Kadav, A.; Durdanovic, I.; Samet, H.; and Graf, H.~P.
\newblock 2016.
\newblock Pruning filters for efficient convnets.
\newblock {\em ICLR}.

\bibitem[\protect\citeauthoryear{Lin \bgroup et al\mbox.\egroup
  }{2019}]{lin2019towards}
Lin, S.; Ji, R.; Yan, C.; Zhang, B.; Cao, L.; Ye, Q.; Huang, F.; and Doermann,
  D.
\newblock 2019.
\newblock Towards optimal structured cnn pruning via generative adversarial
  learning.
\newblock In {\em CVPR}.

\bibitem[\protect\citeauthoryear{Liu \bgroup et al\mbox.\egroup
  }{2017}]{liu2017learning}
Liu, Z.; Li, J.; Shen, Z.; Huang, G.; Yan, S.; and Zhang, C.
\newblock 2017.
\newblock Learning efficient convolutional networks through network slimming.
\newblock In {\em CVPR}.

\bibitem[\protect\citeauthoryear{Luo, Wu, and Lin}{2017}]{luoiccv2017}
Luo, J.-H.; Wu, J.; and Lin, W.
\newblock 2017.
\newblock Thinet: A filter level pruning method for deep neural network
  compression.
\newblock In {\em ICCV}.

\bibitem[\protect\citeauthoryear{Molchanov \bgroup et al\mbox.\egroup
  }{2017}]{molchanov2017taylor}
Molchanov, P.; Tyree, S.; Karras, T.; Aila, T.; and Kautz, J.
\newblock 2017.
\newblock Pruning convolutional neural networks for resource efficient
  inference.
\newblock {\em ICLR}.

\bibitem[\protect\citeauthoryear{Molchanov \bgroup et al\mbox.\egroup
  }{2019}]{molchanov2019importance}
Molchanov, P.; Mallya, A.; Tyree, S.; Frosio, I.; and Kautz, J.
\newblock 2019.
\newblock Importance estimation for neural network pruning.
\newblock In {\em CVPR}.

\bibitem[\protect\citeauthoryear{Rastegari \bgroup et al\mbox.\egroup
  }{2016}]{rastegari2016quantization}
Rastegari, M.; Ordonez, V.; Redmon, J.; and Farhadi, A.
\newblock 2016.
\newblock Xnor-net: Imagenet classification using binary convolutional neural
  networks.
\newblock In {\em ECCV}.

\bibitem[\protect\citeauthoryear{Romero \bgroup et al\mbox.\egroup
  }{2015}]{romero2014fitnets}
Romero, A.; Ballas, N.; Kahou, S.~E.; Chassang, A.; Gatta, C.; and Bengio, Y.
\newblock 2015.
\newblock Fitnets: Hints for thin deep nets.
\newblock {\em ICLR}.

\bibitem[\protect\citeauthoryear{Russakovsky \bgroup et al\mbox.\egroup
  }{2015}]{imagenet}
Russakovsky, O.; Deng, J.; Su, H.; Krause, J.; Satheesh, S.; Ma, S.; Huang, Z.;
  Karpathy, A.; Khosla, A.; Bernstein, M.; et~al.
\newblock 2015.
\newblock Imagenet large scale visual recognition challenge.
\newblock {\em IJCV} 115(3):211--252.

\bibitem[\protect\citeauthoryear{Sandler \bgroup et al\mbox.\egroup
  }{2018}]{sandler2018mobilenetv2}
Sandler, M.; Howard, A.; Zhu, M.; Zhmoginov, A.; and Chen, L.-C.
\newblock 2018.
\newblock Mobilenetv2: Inverted residuals and linear bottlenecks.
\newblock In {\em CVPR}.

\bibitem[\protect\citeauthoryear{Simonyan and
  Zisserman}{2014}]{simonyan2014very}
Simonyan, K., and Zisserman, A.
\newblock 2014.
\newblock Very deep convolutional networks for large-scale image recognition.
\newblock {\em arXiv}.

\bibitem[\protect\citeauthoryear{Soomro, Zamir, and
  Shah}{2012}]{soomro2012ucf101}
Soomro, K.; Zamir, A.~R.; and Shah, M.
\newblock 2012.
\newblock Ucf101: A dataset of 101 human actions classes from videos in the
  wild.

\bibitem[\protect\citeauthoryear{Tran \bgroup et al\mbox.\egroup
  }{2015}]{tran2015learning}
Tran, D.; Bourdev, L.; Fergus, R.; Torresani, L.; and Paluri, M.
\newblock 2015.
\newblock Learning spatiotemporal features with 3d convolutional networks.
\newblock In {\em ICCV}.

\bibitem[\protect\citeauthoryear{Xue, Li, and
  Gong}{2013}]{xue2013factorization}
Xue, J.; Li, J.; and Gong, Y.
\newblock 2013.
\newblock Restructuring of deep neural network acoustic models with singular
  value decomposition.
\newblock In {\em Interspeech}.

\bibitem[\protect\citeauthoryear{Yu \bgroup et al\mbox.\egroup
  }{2018}]{yu2018nisp}
Yu, R.; Li, A.; Chen, C.-F.; Lai, J.-H.; Morariu, V.~I.; Han, X.; Gao, M.; Lin,
  C.-Y.; and Davis, L.~S.
\newblock 2018.
\newblock {NISP}: Pruning networks using neuron importance score propagation.
\newblock In {\em CVPR}.

\bibitem[\protect\citeauthoryear{Zhang \bgroup et al\mbox.\egroup
  }{2016}]{zhang2015factorization}
Zhang, X.; Zou, J.; He, K.; and Sun, J.
\newblock 2016.
\newblock Accelerating very deep convolutional networks for classification and
  detection.
\newblock {\em T-PAMI} 38(10):1943--1955.

\bibitem[\protect\citeauthoryear{Zhang \bgroup et al\mbox.\egroup
  }{2018a}]{zhang2017shufflenet}
Zhang, X.; Zhou, X.; Lin, M.; and Sun, J.
\newblock 2018a.
\newblock Shufflenet: An extremely efficient convolutional neural network for
  mobile devices.
\newblock In {\em CVPR}.

\bibitem[\protect\citeauthoryear{Zhang \bgroup et al\mbox.\egroup
  }{2018b}]{zhang2018three}
Zhang, Y.; Wang, H.; Luo, Y.; and Hu, R.
\newblock 2018b.
\newblock Three dimensional convolutional neural network pruning with
  regularization-based method.
\newblock {\em NeurIPS Workshop}.

\bibitem[\protect\citeauthoryear{Zhao \bgroup et al\mbox.\egroup
  }{2019}]{zhao2019variational}
Zhao, C.; Ni, B.; Zhang, J.; Zhao, Q.; Zhang, W.; and Tian, Q.
\newblock 2019.
\newblock Variational convolutional neural network pruning.
\newblock In {\em CVPR}.

\bibitem[\protect\citeauthoryear{Zhuang \bgroup et al\mbox.\egroup
  }{2018}]{zhuang2018discrimination}
Zhuang, Z.; Tan, M.; Zhuang, B.; Liu, J.; Guo, Y.; Wu, Q.; Huang, J.; and Zhu,
  J.
\newblock 2018.
\newblock Discrimination-aware channel pruning for deep neural networks.
\newblock In {\em NeurIPS}.

\end{thebibliography}
\end{document}